\pdfoutput=1

\documentclass[11pt]{article}

\usepackage[preprint]{acl}

\usepackage{times}
\usepackage{latexsym}
\usepackage{graphicx} 
\usepackage{subfigure}
\usepackage{amssymb,amsthm,dsfont,mathrsfs}
\usepackage{amsmath}
\usepackage{multirow}
\usepackage{mathrsfs}
\usepackage{float}
\usepackage{hyperref}
\usepackage{amsfonts}

\usepackage[linesnumbered, ruled]{algorithm2e}
\usepackage{inconsolata}
\usepackage{subfigure}
\usepackage{amssymb,amsthm,dsfont,mathrsfs}
\usepackage{amsmath}
\usepackage{multirow}
\usepackage{mathrsfs}
\usepackage{float}
\usepackage{hyperref}
\usepackage{adjustbox}
\usepackage{booktabs}
\usepackage{multirow}
\usepackage{lscape}
\usepackage{xcolor}
\usepackage{colortbl}
\usepackage{xspace}
\usepackage{tabularx}
\usepackage[most]{tcolorbox}
\usepackage{longtable}
\usepackage{tikz}
\usetikzlibrary{backgrounds}
\usetikzlibrary{arrows,shapes}
\usetikzlibrary{tikzmark}
\usetikzlibrary{calc}

\usepackage[T1]{fontenc}

\usepackage[utf8]{inputenc}

\usepackage{microtype}

\usepackage{inconsolata}
\newcommand{\ours}{TransferTOD\xspace}

\title{TransferTOD: A Generalizable Chinese Multi-Domain Task-Oriented Dialogue System with Transfer Capabilities} 
 
\author{\normalsize \textbf{Ming Zhang}$^{1}$\thanks{\hspace{1mm} Equal Contribution.} \ \
        \textbf{Caishuang Huang}$^{1*}$\textbf{,} \ \
        \textbf{Yilong Wu}$^{1*}$ \textbf{,} \ \
        \textbf{Shichun Liu}$^{1}$\textbf{,} \ \
        \textbf{Huiyuan Zheng}$^{1}$\textbf{,} \\
        \normalsize \textbf{Yurui Dong}$^{1}$\textbf{,} \ \
        \textbf{Yujiong Shen}$^{1}$\textbf{,} \ \
        \textbf{Shihan Dou}$^{1}$\textbf{,} \ \
        \textbf{Jun Zhao}$^{1}$\textbf{,} \ \
        \textbf{Junjie Ye}$^{1}$\textbf{,} \\
        \normalsize \textbf{Qi Zhang}$^{1,2}$\thanks{\hspace{1mm} Corresponding Author.}\textbf{,} \ \
        \textbf{Tao Gui}$^{2,3}$\textbf{,} \ \
        \textbf{Xuanjing Huang}$^{1,2}$ \\
  {$^1$  \normalsize School of Computer Science, Fudan University} \\
  {$^2$  \normalsize Shanghai Key Laboratory of Intelligent Information Processing, Fudan University}\\
  {$^3$  \normalsize Institute of Modern Languages and Linguistics, Fudan University}\\
  \texttt{\normalsize mingzhang23@m.fudan.edu.cn}\\
  \texttt{\normalsize qz@fudan.edu.cn}\\
}

\begin{document}
\maketitle
\begin{abstract}
Task-oriented dialogue (TOD) systems aim to efficiently handle task-oriented conversations, including information collection. How to utilize TOD accurately, efficiently and effectively for information collection has always been a critical and challenging task. Recent studies have demonstrated that Large Language Models (LLMs) excel in dialogue, instruction generation, and reasoning, and can significantly enhance the performance of TOD through fine-tuning. However, current datasets primarily cater to user-led systems and are limited to predefined specific scenarios and slots, thereby necessitating improvements in the proactiveness, diversity, and capabilities of TOD. In this study, we present a detailed multi-domain task-oriented data construction process for conversations, and a Chinese dialogue dataset generated based on this process, \textbf{TransferTOD}, which authentically simulates human-computer dialogues in 30 popular life service scenarios. Leveraging this dataset, we trained a model using full-parameter fine-tuning called \textbf{TransferTOD-7B}, showcasing notable abilities in slot filling and questioning. Our work has demonstrated its strong generalization capabilities in various downstream scenarios, significantly enhancing both data utilization efficiency and system performance. The data is released in~\url{https://github.com/KongLongGeFDU/TransferTOD}.
\end{abstract}

\section{Introduction}

The Task-Oriented Dialogue (TOD) system is a human-computer interaction system aims to aid users in accomplishing specific tasks or acquiring particular information, which has found extensive use in daily life and commercial applications. At present, TOD systems have displayed the capability to adapt effectively to diverse tasks, domains, and user behaviors. To collect the necessary information, the system must proactively ask questions or guide users to provide the required information for filling specific slots, known as slot filling (SF)~\cite{rosset_spoken_2011}. 

\begin{figure}[!t]
    \includegraphics[width=0.97\linewidth]{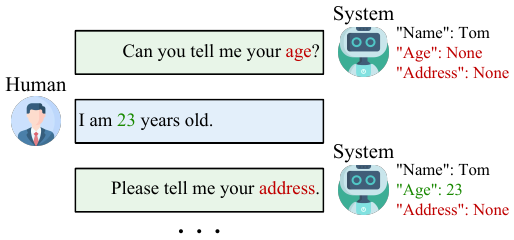}
    \centering
 	\caption{
  In information collection scenaries, the system will ask the user for one or more slot values that are `none', and then identify and update the corresponding field(s) based on the user's response until all slots are filled. For instance, if the system inquires about the user's age, and the user replies with `23', the system will update the slot form `none' to 23.
  }
	\label{fig:intro}
\end{figure}

Although various approaches has been explored to enhance data efficiency (e.g. transfer learning and fine-tuning)~\cite{devlin_bert_2019,liu_roberta_2019,henderson_convert_2020} , traditional SF methods still rely on expert-labeled data~\cite{fuisz_improved_2022}, which is costly and inefficient, and are limited to predetermined scenarios and time periods. Moreover, existing slot filling datasets mostly serve user-driven dialogue systems, typically constrained to specific scenarios and slots predetermined for constructing systems that respond to user inquiries and requests~\cite{wen_network-based_2017, budzianowski_multiwoz_2020, zhu_crosswoz_2020}. In system-driven information collection scenarios, the slots and values that need to be filled are often unfamiliar with the training set, leading to a significant decline in the accuracy of slot filling. Additionally, the system struggles to ask precise questions based on the filled slots to guide user responses, thereby completing the next slot-filling action. Traditional TOD systems face significant challenges in this task.

Recently, Large Language Models (LLMs) have exhibited promising performance in dialogue participation, instruction generation, and zero-shot reasoning~\cite{zhang_sgp-tod_2023}, which brought new ideas to solving the above problems. Research has confirmed that fine-tuned LLMs on dialogue corpora of different sizes can achieve enhanced performance across diverse tasks, domains, and even languages~\cite{DBLP:journals/corr/abs-2103-10360, touvron2023llama}. Hence, we can use LLMs to drive TOD and solve some problems that were difficult to solve in the small model era.

In this paper, inspired by real-world customer information collection scenarios as shown in the Figure~\ref{fig:intro}, we proposed a dataset called TransferTOD. This multi-domain, task-oriented Chinese dialogue dataset is designed to handle complex and diverse tasks, simulating realistic conversation scenarios. To account for potential human errors and practical applications in the Chinese context, we implemented a four-step dataset construction pipeline, including noise injection and language polishing. This process resulted in a robust dataset containing 35965 turns and 5460 dialogues across 30 life service scenarios, offering a valuable resource for research. Using this dataset, we trained the TransferTOD-7B model with appropriate base models and fine-tuning methods. The model can proactively ask users for missing information, accurately fill slots, and efficiently guide fluent response generation. Our evaluations of slot-filling ability and semantic accuracy demonstrated that the dataset significantly improves model's performance by handling noise, increasing question diversity, and optimizing language fluency.

Summarizing, the principal contributions of our paper are as follows:

1. We propose a comprehensive four-step pipeline with high generalizability, allowing researchers to create new TOD datasets across different languages or in multilingual contexts through similar methods.

2. We construct a new dataset called TransferTOD for task-oriented dialogue generation in various lifestyle service scenarios, containing 5460 dialogues across 30 scenarios. Ablation experiments have demonstrated that this dataset exhibits good noise resistance, diversity, and fluency.

3. We utilize TransferTOD to fine-tune the TransferTOD-7B model, achieving superior slot-filling and questioning capabilities compared to GPT-4. With appropriate secondary fine-tuning, our model demonstrates superior performance in out-of-domain test compared to GPT-3.5-Turbo fine-tuned with an equivalent amount of data.

\section{Related Work}
\paragraph{Task-oriented Dialogue Datasets} The performance of intelligent dialogue systems is profoundly influenced by the quality of the dialogue datasets, making dataset construction an active research area. Initial generations of task-oriented dialogue datasets often focused on a single task or even a single scenario, ATIS~\cite{hemphill_atis_1990}, DSTC2~\cite{henderson_second_2014}, WOZ2.0~\cite{wen_network-based_2017}, etc. included. 


The emergence of these databases not only enhanced the conversational fluency of conversational agents but also made task completion through natural dialogues between machines and humans possible. Considering that user dialogues often involve domain transitions, datasets Multi-WoZ~\cite{budzianowski_multiwoz_2020}, CrossWoZ~\cite{zhu_crosswoz_2020} etc. encompassing more scenes and larger volumes of data were subsequently proposed. 


However, these dialogues are user-led discussions on relevant topics, requiring a user to pose questions or set tasks for the dialogue agent to respond accordingly.

\paragraph{TOD System Enhancement Methodology}
Enhancing the performance and data utilization of TOD systems and strengthening their ability to understand specific tasks expressed by users remain hot research topics. To complete tasks and improve accuracy, Li et al.~\cite{li_end--end_2018} proposed an end-to-end neural dialogue system based on reinforcement learning. TOD gradually started to realize across tasks~\cite{peng_composite_2017}, domains \cite{hakkani-tur_multi-domain_2016}, and even languages~\cite{wang_kddres_2021}. TOD-BERT \cite{wu_tod-bert_2020}, MinTL~\cite{lin_mintl_2020}, Soloist \cite{peng_soloist_2021}, etc. has been successively proposed improving the success rate of tasks. However, as task complexity increases, these methods still rely heavily on large-scale datasets and lack competitiveness in handling noise robustness.

\paragraph{LLM-based TOD System}
Existing research \cite{brown_language_2020,chowdhery_palm_2022,chen_evaluating_2021,openai_gpt-4_2023} has demonstrated LLMs' exceptional capabilities in natural language understanding, zero-shot reasoning, and command generation. With their advent and deep utilization, dialogue systems have entered the LLM-based era \cite{wang_survey_2023}. Utilizing LLMs, many dialogue tasks have achieved significant breakthroughs. On one hand, through internal dialogues with users, systems can be equipped with human-like perception and reasoning abilities, including intent classification, semantic parsing, dialogue state tracking, and reply generation. On the other hand, the integration of external information sources, such as specific databases, memory knowledge sources, the internet, etc., ensures the system provides the latest, rich, accurate, personalized, and necessary information to complete tasks.

\section{TransferTOD}
\subsection{Dataset}
\label{sec:DataSet}

\begin{table}[t]
\centering
\resizebox{0.95\linewidth}{!}{
\begin{tabular}{lccc}
\toprule
 & Train & ID Test & OOD Test\\
\midrule
\# Domain    & 27   & 27    & 3\\
\# Slot      & 188  & 188   & 27\\
\# Dialogue  & 4320 & 540  & 600\\
\# Turns     & 28680 &  3585  &3700\\
\# Slots / Dialogue & 10.3 & 10.3  & 9.7\\
\# Tokens / Turn & 66.4 & 66.4  &76.8\\
\bottomrule
\end{tabular}
}
\caption{Overall statistics of \ours. ID Test means In-Domain test and OOD Test means Out-of-Domain test. The domains of the test set are Water-Delivery, Sanitation, and Courier.}
\label{table:clean-dataset}
\end{table}

TransferTOD aims to construct a cross-disciplinary task-oriented customer information collection multi-turn dialogue dataset, encompassing tasks such as goal-oriented questioning, dialogue state maintenance, information collection, and parsing. Existing task-oriented Wizard of Oz (WoZ) datasets are typically user-driven systems with relatively single domains. Departing from scenarios in the real world where customer information collection dialogue may occur, we have curated dialogues spanning 30 different domains. We have enhanced the data in terms of robustness, diversity, and fluency, ensuring that the data closely mirrors real-world situations.

\begin{figure*}[t]
    \includegraphics[width=0.94\linewidth]{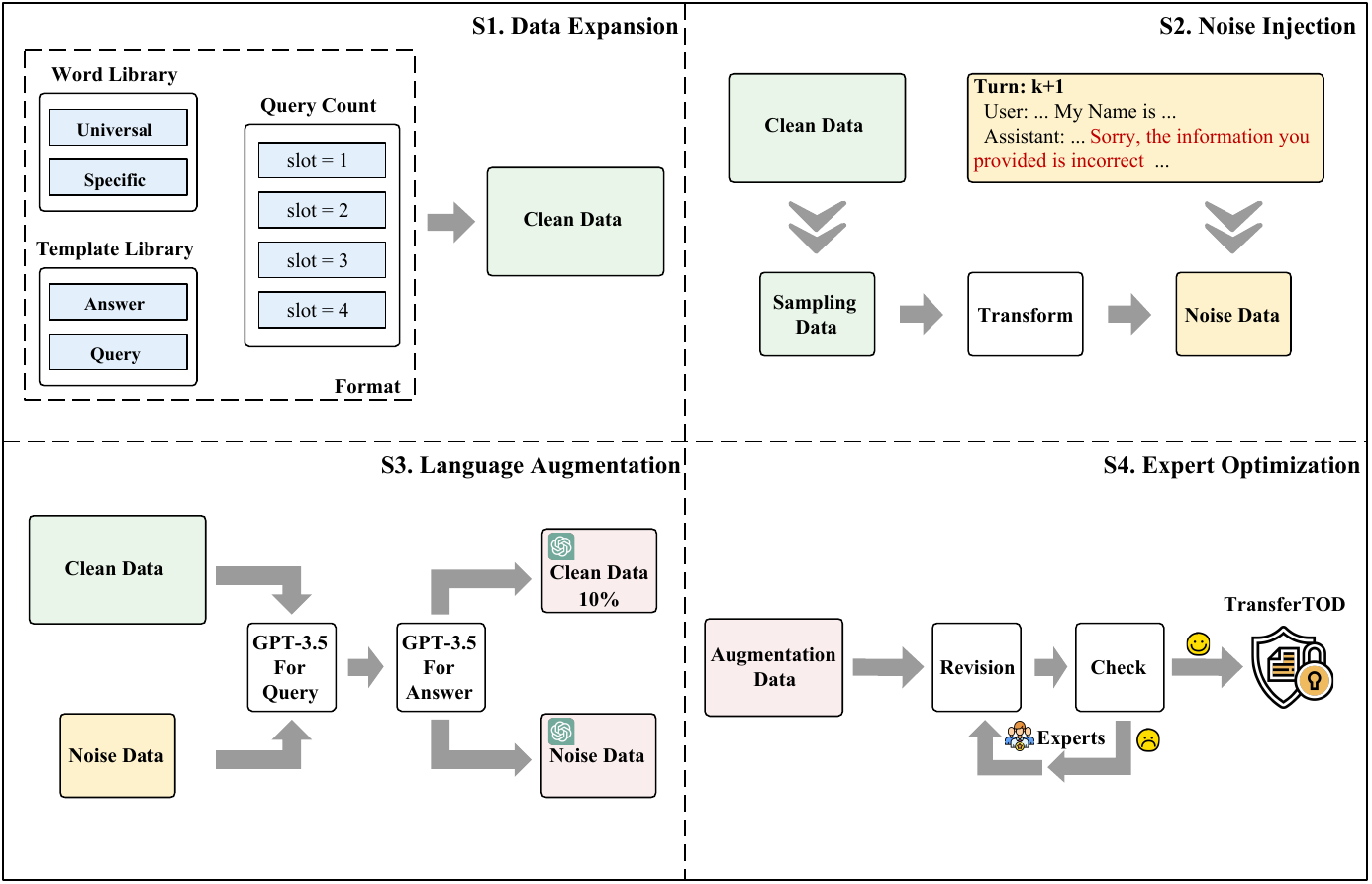}
    \centering
 	\caption{
  Our dataset development comprises four phases. Initially, we create specific scenarios, develop the corresponding questions and answers, and generate data for slots 1 to 4 using rule-based methods. In the second phase, we introduce noise into a subset of this data to simulate inaccuracies needing correction by customer service, prompting a re-query in the next interaction. The third phase diversifies the dataset by rephrasing both clean and noise data via GPT-3.5. In the final stage, expert professionals refine the input to achieve a high level of naturalness in customer service interactions, ensuring that the inquiries exhibit a seamless and fluent conversational flow.
  }
	\label{fig:dataset-build}
\end{figure*}

Figure \ref{fig:dataset-build} illustrates the 4 steps of data collection and processing: 1. Original slot construction and dialogue generation; 2. Introduction of perturbed data; 3. GPT-enhanced dialogue diversity; 4. Manual refinement of dialogue content for fluency. Overall statistics of \ours are shown in Table \ref{table:clean-dataset}.

\subsubsection{ Field Selection and Slot Collection}
We crawl the most popular 30 life service offerings from local lifestyle applications (such as Meituan and Yelp) to construct the domain for our dialogue system. Specifically, we analyzed the submitted forms of each service, abstracting the information that the system would require users to provide as slots. 

After constructing the slots, we built a corpus containing all possible values for each slot. For string-type slots, we adopted a method of collecting publicly available information from the internet and generating rules. 
During the collection process, we kept the information for each slot separate, ensuring that no real personal information was involved. For number-type data, we described its range and distribution, generating it in real-time during the dialogue construction process. 

Human experts\footnote{Details of the human experts are shown in the appendix \ref{app:Experts in Constructing Datasets}} manually created a set of high-quality dialogues as test data across 30 domains; three of these domains were selected for constructing an out-of-domain test set due to their minimal overlap in slots with the other domains. The remaining data is used as the in-domain test set. For the training dataset, the following steps will be undertaken to generate it on a large scale.

\subsubsection{Dialog Construction}
\label{sec:Dialog construction}
Based on existing slot type descriptions and vocabularies, we have implemented the first version of a dialogue dataset using a script-generated approach. Specifically, we constructed a template library for each domain \footnote{Examples of our templates are shown in the appendix \ref {app:templated}}.
Each dialogue round consists of a user response, a system question, or a summary, forming the values before and after the dialogue state changes.

 For the number of slots $k$ that could potentially be extracted in a single dialogue, we experimented with four scenarios: $k=1, 2, 3, 4$. Specifically,  when $k=3$, the system simultaneously asks the user for information on 3 slots, and the user needs to respond to these three corresponding aspects. 
 The statistical information of the original dialogue data is detailed in Table \ref{table:clean-dataset}.
The dataset obtained after this step is \textbf{TransferTOD-v1}.


\subsubsection{Noisy Data Construction}
In real-world scenarios, users may provide information that does not conform to standards or common sense. Therefore, a comprehensive dialogue system should possess the capability to scrutinize the responses provided by users and, when necessary, seek clarification to obtain accurate information. To address this, a portion of the data is delineated to incorporate rounds of interaction specifically designed to handle incorrect responses from users. 

There are two types of data disturbances: 1. Non-responsive answers, where the content of the user's reply significantly deviates from the system's query. This dialogue alteration is achieved by replacing the user's response with an irrelevant answer; 2. Illogical responses, where the user's reply may contradict basic common sense. This data segment necessitates the introduction of non-factual content into the slot value lexicon to accommodate such instances.

During rounds with erroneous responses, the system will identify the user's mistake, repeat the original question, and maintain the dialogue state without updating it. We constructed 3013 noise dialogue data, with each dialogue containing at least one of the aforementioned errors, where the first type of error represented more than 90\% of the cases. The dataset obtained after this step is \textbf{TransferTOD-v2}. Data examples are shown in appendix \ref{data_examples}.

\subsubsection{ Dialogue Diversity and Fluency Polish}
Dialogue data generated by static script schemes exhibit a shortfall in the diversity of questioning and answering modes. Each slot is confined to merely 5-6 variations of queries and responses, which fails to mirror the spectrum of linguistic preferences encountered in real-life scenarios. Consequently, we have leveraged the GPT-3.5 model to reformulate the texts of questions and answers, ensuring fidelity to the original intents while adjusting the temperature coefficient to 0.5 for an enriched array of textual content. The dataset obtained after this step is \textbf{TransferTOD-v3}.

Furthermore, we have refined the fluidity of dialogues that encompass inquiries about multiple slots within a singular exchange. Initially, dialogue data were essentially composed of simplistic amalgamations of disjointed questions or responses, not aligning with conventional spoken habits. Through the application of GPT-4 for sentence amalgamation and enhancement of coherence, coupled with rule-based scrutiny to pinpoint instances of fragmented sentences, we have engaged human annotators for the revision of overlooked or non-compliant sentences, thereby assuring dialogue smoothness. The dataset obtained after this step is \textbf{TransferTOD-v4}, which is our final dataset. Prompts are shown in appendix \ref{prompts}.

\subsection{Models}
Upon acquiring the TransferTOD dataset, we opted for the Baichuan2-7B-Base \cite{baichuan2023baichuan2} as the foundational model for fine-tuning. During the model training process, we employed two methods: full-parameter fine-tuning \cite{Zeng2023AgentTuningEG} and LoRA (Low-Rank Adaptation) fine-tuning \cite{hu2021lora}.

\subsubsection{Supervised Fine-tuning}
To equip the model with basic conversational abilities, we initially combined the training subset of the TransferTOD dataset with the general Chinese conversational dataset BELLE \cite{belle2023exploring} in equal proportions to construct the SFT (Supervised Fine-Tuning) dataset. This dataset was utilized for full-parameter fine-tuning of the Baichuan2-7B-Base model to derive the TransferTOD-7B model.

\subsubsection{Secondary Fine-tuning}
Following the development of our TransferTOD-7B model, we aimed for our model to achieve commendable performance in specific downstream tasks, necessitating that our model possesses superior generalization capabilities. In three external domain test sets, we adopted a limited-sample secondary fine-tuning approach to further enhance the accuracy of TransferTOD-7B in external domain test sets. Research \cite{Sun2023ACS} indicates that compared to full-parameter fine-tuning, LoRA fine-tuning achieves better generalization. Consequently, we employed LoRA fine-tuning for secondary fine-tuning. Experimental evidence demonstrates the effectiveness of our methodology in scenarios where data availability for downstream tasks is significantly constrained.

\section{Experiment}

In this section, we detail the experiments conducted. The primary experiments were carried out on the test set of TransferTOD. Additionally, we conducted ablation studies on the dataset construction phase, as well as supplementary experiments to further investigate the effects of secondary fine-tuning.

\subsection{Experimental Setup}

For the in-domain test, we evaluated various methods known for their effectiveness in slot extraction. Traditional TOD systems divide the task into several modules \cite{zhu_crosswoz_2020}, each managed by a distinct model, forming a system pipeline. However, LLMs can reduce the reliance on task decomposition, thereby allowing us to directly evaluate the core competency of slot filling through information extraction.

For the out-of-domain test, a model's ability to adapt and generalize is paramount. Consequently, our initial evaluation centered on a selection of open-source LLMs with parameter counts comparable to our base model (7 billion), all of which demonstrated strong performance in Chinese benchmarks. To further enhance our analysis, we incorporated two powerful, near-source models from OpenAI.

\subsubsection{Baseline}
For the in-domain test, we select 4 models as baseline: BertNLU~\cite{zhu_crosswoz_2020}, SoftLexicon (LSTM)~\cite{ruotian2020simplify}, LEBERT+CRF~\cite{liu-etal-2021-lexicon} and W2NER~\cite{li2022unified}.\\
For the out-of-domain test, we select 6 Large Language Models as baseline: {Baichuan2}~\cite{baichuan2023baichuan2}, {ChatGLM3}~\cite{du2022glm, zeng2022glm}, {Qwen}~\cite{qwen}, {Yi}\footnote{\url{https://github.com/01-ai/Yi}}, {GPT-3.5-Turbo}\footnote{\url{https://platform.openai.com/docs/models/gpt-3-5}}, {GPT-4}~\cite{openai_gpt-4_2023}. Please refer to the appendix \ref{app:baselines} for details.

\subsubsection{Implementation Detail}


For evaluating the slot filling capability, we have annotated user utterances with BIO tags and trained 4 models for the in-domain test. A detailed system prompt was designed when inferencing with those LLMs in out-of-domain test. Please refer to the appendix \ref{app:Implementation Details} for details.

\subsubsection{Evaluation Metrics}
For the out-of-domain test, we assess the model's capabilities in two main aspects: slot filling ability and semantic accuracy during the phase of guiding user responses. To evaluate the slot filling ability, we employ F1 and Joint Accuracy, which are widely used in the TOD systems for slot extraction tasks. To evaluate the semantic accuracy of model-generated questions, we use a manual evaluation approach. Please refer to the appendix \ref{app:metrics} for details.

It is worth noting that we use the Dialog Act F1 as the evaluation metric in this context. While the Dialog Act F1 and SlotF1 metrics appear to be calculated in the same way, they differ slightly in essence. For a detailed explanation of these subtle differences, please also refer to the appendix \ref{app:metrics}.


\subsection{Results on TransferTOD}
This section shows the results of our main experiment. 
\subsubsection{Results on In-Domain Test}
Table \ref{tab:id_result} presents the results of the in-domain test. Compared with traditional methodologies including W2NER, the State-Of-The-Art model in several ChineseNER tasks, our model significantly outperforms others on the in-domain test set in terms of the Dialogue Act F1 Score. This underscores the exceptional slot-filling accuracy of our model within domain-specific data.

\begin{table}
    \centering
    \resizebox{\linewidth}{!}{
    \begin{tabular}{c c}
    \toprule
    \textbf{Model} & \textbf{Dialogue Act F1(\%)}\\
    \midrule
    BertNLU~\cite{zhu_crosswoz_2020}    & 79.32 \\
    SoftLexicon (LSTM)~\cite{ruotian2020simplify}  & 77.12 \\
    LEBERT+CRF~\cite{liu-etal-2021-lexicon}  & 79.72 \\
    W2NER~\cite{li2022unified}      & 78.45 \\
    \textbf{TransferTOD-7B}   & \textbf{93.64} \\
    \bottomrule
    \end{tabular}
    }
    \caption{Results on in-domain test: The Dialogue Act F1 Score of each model, showing the accuracy of predicting the right dialogue acts from user utterance.}
    \label{tab:id_result}
\end{table}

\subsubsection{Results on Out-of-Domain Test}

\begin{table*}
    \centering
    \resizebox{\linewidth}{!}{
    \begin{tabular}{ c | c c c c c c c c}
    \toprule
     \multicolumn{2}{c}{\textbf{Model}}& \textbf{Scenario} & \textbf{JointAcc(\%)} & \textbf{SlotF1(\%)} & \textbf{AVG.JointAcc(\%)} & \textbf{AVG.SlotF1(\%)} & \textbf{Ask\_Acc} & \textbf{Ask\_Flu}\\
    \midrule
      \multirow{15}{0.15\linewidth}{\textbf{Open-Source Model}}&\multirow{3}{*}{\textbf{Baichuan2-7B-Chat}} & Water-Delivery & 15.26& 41.83& \multirow{3}{*}{19.44}& \multirow{3}{*}{44.93}& \multirow{3}{*}{27.50}&\multirow{3}{*}{25.50}\\
      && Sanitation& 24.29& 46.17& & & &\\
      && Courier& 18.77& 46.79& & & &\\
     \cmidrule{2-9}
      &\multirow{3}{*}{\textbf{BlueLM-7B-Chat}} & Water-Delivery & 0.80& 3.17& \multirow{3}{*}{0.27}& \multirow{3}{*}{1.06}& \multirow{3}{*}{3.50}&\multirow{3}{*}{0.17}\\
      && Sanitation& 0.00& 0.02& & & &\\
      && Courier& 0.00& 0.00& & & &\\
     \cmidrule{2-9}
      &\multirow{3}{*}{\textbf{Chatglm3-6B}} & Water-Delivery & 4.47& 23.03& \multirow{3}{*}{4.11}& \multirow{3}{*}{21.14}& \multirow{3}{*}{25.67}&\multirow{3}{*}{52.67}\\
      && Sanitation& 4.48& 23.99& & & &\\
      && Courier& 3.38& 16.41& & & &\\
     \cmidrule{2-9}
      &\multirow{3}{*}{\textbf{Qwen-7B-Chat}} & Water-Delivery & 17.01& 38.13& \multirow{3}{*}{17.14}& \multirow{3}{*}{38.47}& \multirow{3}{*}{28.67}&\multirow{3}{*}{30.67}\\
      && Sanitation& 16.57& 33.45& & & &\\
      && Courier& 17.85& 43.83& & & &\\
     \cmidrule{2-9}
      &\multirow{3}{*}{\textbf{Yi-6B-Chat}} & Water-Delivery & 1.04& 5.87& \multirow{3}{*}{1.22}& \multirow{3}{*}{4.59}& \multirow{3}{*}{22.33}&\multirow{3}{*}{52.83}\\
      && Sanitation& 0.76& 2.92& & & &\\
      && Courier& 1.85& 4.98& & & &\\
      \midrule
      \multirow{6}{0.15\linewidth}{\textbf{Close-Source Model}}&\multirow{3}{*}{\textbf{GPT-3.5-Turbo}} & Water-Delivery & 41.69& 74.64& \multirow{3}{*}{35.71}& \multirow{3}{*}{69.44}& \multirow{3}{*}{72.17}&\multirow{3}{*}{77.67}\\
      && Sanitation& 31.43& 65.44& & & &\\
      && Courier& 34.00& 68.24& & & &\\
      \cmidrule{2-9}
      &\multirow{3}{*}{\textbf{GPT-4-1106-Preview}} & Water-Delivery & 42.01& 74.21& \multirow{3}{*}{41.68}& \multirow{3}{*}{70.91}& \multirow{3}{*}{\textbf{90.00}}&\multirow{3}{*}{72.33}\\
      && Sanitation& 40.19& 68.32& & & &\\
      && Courier& 42.85& 70.18& & & &\\
      \midrule
      \multicolumn{2}{c}{\multirow{3}{*}{\textbf{TransferTOD-7B}}}& Water-Delivery & \textbf{73.16}& \textbf{96.61}& \multirow{3}{*}{\textbf{75.09}}& \multirow{3}{*}{\textbf{96.20}}& \multirow{3}{*}{75.00}&\multirow{3}{*}{\textbf{84.00}}\\
      \multicolumn{2}{c}{}& Sanitation& \textbf{84.09}& \textbf{97.43}& & & &\\
      \multicolumn{2}{c}{}& Courier& \textbf{68.00}& \textbf{94.57}& & & &\\
     \bottomrule
    \end{tabular}
    }
    
    \caption{Results on out-of-domain test: The Joint Accuracy and Slot F1 Score of each model, showing the accuracy of predicting the right dialogue state and slot-value pairs respectively.}
    \label{tab:ood_result}
\end{table*}

Table \ref{tab:ood_result} showcases the results for the out-of-domain test set. The findings affirm that the average joint accuracy of TransferTOD-7B reached 75.09\%, with a Slot F1 Score of 96.20\%, surpassing other large-scale models, including the most advanced GPT-4, which only achieved a joint accuracy of 41.68\%. In terms of query selection, GPT-4 leads the performance compared to other open-source models. TransferTOD's performance in this aspect scored 75, trailing just behind GPT-4. However, TransferTOD surpassed other models in terms of the fluency of queries. Besides, we have conducted a further experiment to compare our TransferTOD-7B to both open-source and close-source model with In-Context Learning 5-shot setting, reducing the probability of poor score caused by wrong format, the results are presented in Table \ref{tab:ood_result_icl}, showing our TransferTOD's superior performance.

The experimental results validate that our TransferTOD model possesses robust generalization capabilities, achieving nearly 80\% accuracy in specific downstream tasks. With appropriate secondary fine-tuning, the overall score can be further enhanced.

\subsection{Secondary Fine-Tuning Study}
\subsubsection{Secondary Fine-Tuning}
In this section, we primarily discuss our experiments on performing secondary fine-tuning on TransferTOD-7B. The objective was to simulate enhancing our model's slot filling and question-asking capabilities in external scenarios using a small subset of downstream scenario data. We fine-tuned GPT-3.5-Turbo\footnote{\url{https://platform.openai.com/docs/guides/fine-tuning}} as our baseline and conducted fine-tuning with 50, 100 and 200 pieces of data across three out-of-domain scenarios, respectively. The remaining data served as the test set for this experiment.

In the third scenario (Courier), we undertook multiple experiments employing various fine-tuning strategies, such as adding BELLE \cite{belle2023exploring} dataset, incorporating in-domain data, and upsampling out-of-domain scenario data. This research aimed to identify methods that could further enhance the TransferTOD-7B model's slot filling capabilities.
\subsubsection{Result}

\begin{table*}
    \centering
    \resizebox{\linewidth}{!}{
    \begin{tabular}{c | c c c c c c}
    \toprule
    \multirow{2}{*}{\textbf{Scenario}} & \multirow{2}{*}{\textbf{Model}} & \multirow{2}{*}{\textbf{Num.ScenarioData}} & \multicolumn{2}{c}{\textbf{Num.OTD}} & \multirow{2}{*}{\textbf{JointAcc(\%)}} & \multirow{2}{*}{\textbf{SlotF1(\%)}}\\
    & & & \textbf{TransferTOD} & \textbf{Belle} & & \\
    \midrule
    \multirow{4}{*}{\textbf{Water-Delivery}}& \multirow{2}{*}{\textbf{GPT-3.5-Turbo}}& 0& /& /& 41.69& 74.64\\
    & &  50& /& /& 71.49& 93.53\\
    \cmidrule{2-7}
    & \multirow{2}{*}{\textbf{TransferTOD-7B}} &  0 & 0 & 0 & 73.16 & 96.61 \\
    & &  50& 0& 0& \textbf{73.48} & \textbf{96.64} \\
    \midrule
    \multirow{4}{*}{\textbf{Sanitation}}& \multirow{2}{*}{\textbf{GPT-3.5-Turbo}}& 0& /& /& 31.43& 65.44\\
    & &  100& /& /& 78.48& 95.78\\
    \cmidrule{2-7}
    & \multirow{2}{*}{\textbf{TransferTOD-7B}} &  0 & 0 & 0 & 84.09 & 97.43\\
    & &  100& 0& 0& \textbf{84.95} & \textbf{97.54}\\
    \midrule
    \multirow{7}{*}{\textbf{Courier}}& \multirow{2}{*}{\textbf{GPT-3.5-Turbo}}& 0& /& /& 34.00& 68.24\\
    & &  200& /& /& \textbf{78.54}& 91.01\\
    \cmidrule{2-7}
    & \multirow{5}{*}{\textbf{TransferTOD-7B}} &  0 & 0 & 0 & 68.00 & 94.57\\
    & &  200   &    0&    0& 69.08 & 94.83\\
    & &  200$\times$4 & 8000&    0& 69.62 & 95.13\\
    & &  200$\times$4 & 8000& 8000& 68.38 & 94.81\\
    & &  200$\times$8 & 8000&    0& 70.15 & \textbf{95.19}\\
    \bottomrule
    \end{tabular}
    }

    \caption{Result of Secondary Fine-Tuning on out-of-domain test: The Joint Accuracy and Slot F1 Score of each model, showing the accuracy of predicting the right dialogue state and slot-value pairs respectively. "OTD" stands for Original Train Data which is used in fine-tuning the TransferTOD-7B. "200$\times$4" in Num.ScenarioData represents that we took 200 ScenarioData and repeated it four times.}
    \label{tab:secondary_ft_result}
\end{table*}
Table \ref{tab:secondary_ft_result} shows the results of fine-tuning GPT3.5 and TransferTOD-7B in scenarios. The secondary fine-tuning can improve the model's out-of-domain capability. After fine-tuning, TransferTOD-7B still outperform GPT-3.5 (especially SlotF1) in most cases.

\subsection{Ablation Studies}
Based on the TransferTOD-v1, v2, v3, and v4 mentioned in \ref{sec:DataSet}, we trained models TransferTOD-7B-v1 to v4 individually. To ascertain the efficacy and trustworthiness of our data construction methodologies, we rigorously assessed their performance in terms of robustness, diversity and fluency. The method we employed, which combines GPT-based assessment with expert review, is a widely adopted approach for evaluating the language fluency of models \cite{10.1145/3641289, DBLP:conf/aaai/ZhangZYLSG0H24}. For details on the GPT assessment instructions and the expert review process, please refer to the appendix \ref{app:Prompt GPT-4 to evaluate the results} and \ref{app:Experts in Ablation Experiment}.

\paragraph{Noise Injection} To strengthen the model's resilience to noise, we augmented the standard dataset with a controlled amount of noisy data and trained the TransferTOD-7B-v2 model on it. As shown in Table \ref{tab:noise_result}, the improvement in joint accuracy substantiates the hypothesis that incorporating noisy data indeed strengthens the model's resistance to noise.

\begin{table}
    \centering
    \resizebox{\linewidth}{!}
    {
    \begin{tabular}{c c c}
    \toprule
    \textbf{Model}  &  \textbf{JointAcc(\%)}  &  \textbf{SlotF1(\%)}\\
    \midrule
    TransferTOD-7B-v1  & 11.91 &  80.53 \\
    TransferTOD-7B-v2  & 55.50 &  90.24 \\
    \bottomrule
    \end{tabular}
    }
    \caption{Results on Noise Injection: The Joint Accuracy and Slot F1 Score of each model, showing the accuracy of predicting the right dialogue state and slot-value pairs respectively.}
    \label{tab:noise_result}
\end{table}

\begin{figure}[!]
    \includegraphics[width=0.97\linewidth]{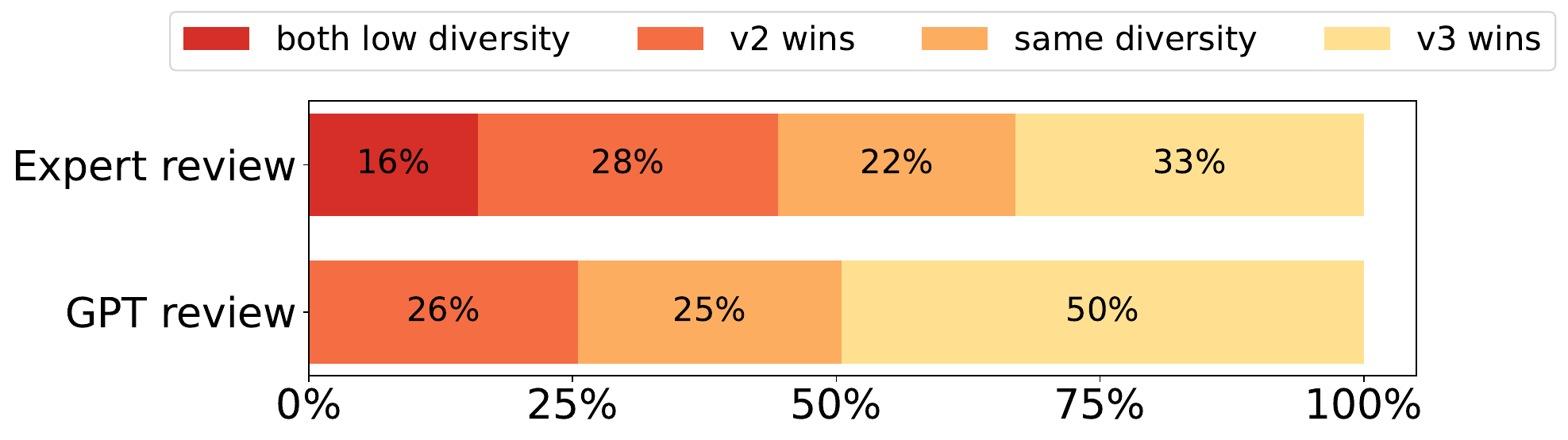}
    \centering
 	\caption{Results on Language Augmentation: Different color blocks  represent the winning rate of two version models in terms of diversity comparison.
  }
	\label{fig:diversity}
\end{figure}

\paragraph{Language Augmentation} To enhance the diversity of model interrogation techniques, we expanded our dataset by leveraging GPT, followed by a comprehensive assessment of the diversity in the questions generated by the newly developed models. Evaluators were provided with four assessment options: model A exhibits superior diversity, model B exhibits superior diversity, both models demonstrate comparable diversity, or neither model exhibits satisfactory diversity. The assessment results collected are shown in the Figure \ref{fig:diversity}. Both the outcomes of expert review and GPT review affirm that v3 model surpasses v2 model in linguistic diversity.


\paragraph{Fluency Enhancement} To enhance the fluency of the model's inquiries, we manually revised the dataset and employed a hierarchical scoring system to evaluate the models' query smoothness. The findings, as delineated in Table \ref{table:fluency}, underwent normalization to a 100-point scale, unequivocally demonstrate an improvement in the model's questioning fluency. The calculation of fluency score is given by equation \ref{eq:fluency_score}.The experimental results demonstrate that the v4 model outperforms v3 model in both the high score rate on the GPT-based review and the expert review, as well as in terms of fluency score. Thus, our method effectively models query fluency.


\begin{table}
\centering
\resizebox{\linewidth}{!}{
\begin{tabular}{lcccc}
\hline
\textbf{Review Type} & \textbf{0-1 points(\%)} & \textbf{2-3 points(\%)} & \textbf{Ask\_Flu} \\ \hline
v3's GPT review      & 4.50   & 95.50  & 95.33 \\
v4's GPT review      & 2.00   & 98.00  & 97.83 \\
v3's Expert review   & 21.50  & 78.50  & 70.50 \\
v4's Expert review   & 17.50  & 82.50  & 75.00 \\ \hline
\end{tabular}
}

\caption{Results on Fluency Enhancement: The table shows the proportion of high and low scoring questions in GPT and expert ratings, as well as the corresponding total Ask\_Flu score.}
\label{table:fluency}
\end{table}

\section{Conclusion}
Empirical evidence substantiates that our TransferTOD dataset possesses substantial noise resilience and superior linguistic performance. Utilizing this dataset for supervised fine-tuning, the resultant model, designated TransferTOD-7B, attains a joint accuracy of 75.09\% in out-of-domain evaluations, accompanied by a Slot F1  of 96.20\%. When it comes to question-asking ability, the accuracy of TransferTOD-7B is only slightly inferior to GPT-4, whereas its fluency in generating questions surpasses all other models we tested.

Furthermore, our findings suggest that appropriate secondary fine-tuning of the TransferTOD-7B model can further enhance its generalization capabilities. By employing a small portion of the out-of-domain test set for secondary fine-tuning, the resulting model surpasses the performance of GPT-3.5-Turbo, which was fine-tuned with an equivalent amount of data.

In summary, we have proposed a highly versatile data construction process that enhances the quality of task-oriented dialogue data for information collection tasks. The models fine-tuned with this data exhibit strong generalization capabilities, performing well in out-of-domain scenarios.

\section*{Limitations}
Our research presents a comprehensive set of experiments, yet it is not without limitations. One significant constraint stems from our dataset being primarily in Chinese, which precluded the testing of other major English-language open-source models due to their suboptimal performance on tasks in Chinese. Additionally, our assessment of question-asking accuracy employed manual evaluation methods, potentially introducing a degree of subjectivity despite our efforts to minimize such bias.

\section*{Acknowledgments}
The authors wish to thank the anonymous reviewers for their helpful comments. This work was partially funded by National Natural Science Foundation of China (No.62076069,62206057,61976056), Shanghai Rising-Star Program (23QA1400200), and Natural Science Foundation of Shanghai (23ZR1403500).

\bibliography{anthology,custom}

\appendix
\section{Experimental Details}
\label{app:Experimental Details}
\subsection{Baselines}
\label{app:baselines}
For the in-domain test, we select 4 models as baseline:
\paragraph{BertNLU}~\cite{zhu_crosswoz_2020}
is a BERT-based NLU model, initialized with Chinese pre-trained BERT and fine-tuned on tagged training data. For the input word embeddings, utilize MLP to generate BIO-tagged outputs.
\paragraph{SoftLexicon (LSTM)}~\cite{ruotian2020simplify} is an effective method for incorporating the word lexicon into the character by categorizing the matched words, condensing the word sets and combining them with character representation.
\paragraph{LEBERT+CRF}~\cite{liu-etal-2021-lexicon}
Lexicon Enhanced BERT for Chinese sequence labeling, utilizing a Lexicon adapter layer to integrate external lexicon knowledge into BERT layers.
\paragraph{W2NER}~\cite{li2022unified}
is a modeling method of neighboring relations between entity words with Next-Neighboring-Word and Tail-Head-Word-* relations.

For the out-of-domain test, we select 6 Large Language Models as the baseline:
\paragraph{Baichuan2}~\cite{baichuan2023baichuan2} is an open-sourced large language model trained on 2.6 trillion tokens, achieving top performance in various Chinese and multilingual benchmarks. We utilized Baichuan2-7B-chat for our experiments.

\paragraph{ChatGLM3}~\cite{du2022glm, zeng2022glm} 
is Jointly developed by Zhipu AI and Tsinghua University, is the strongest in its class for datasets across multiple disciplines, supporting complex tasks like function calls and code interpretation. We utilized ChatGLM3-6B for our experiments.

\paragraph{Qwen}~\cite{qwen}
Trained on 3 trillion tokens across multiple languages, Qwen models show competitive performance, excelling in tasks like chatting, text generation, and information extraction. We utilized Qwen-7B-chat for our experiments.

\paragraph{Yi}\footnote{\url{https://github.com/01-ai/Yi}}
A powerful bilingual model, demonstrating significant potential in language cognition and reasoning, ranking highly on the SuperCLUE leaderboard and surpassing other large models in Chinese language proficiency. We utilized Yi-6B-chat for our experiments.

\paragraph{BlueLM}~\cite{2023bluelm} is
a large-scale model from vivo AI Global Research Institute, trained on a 2.6 trillion token corpus, showing leading results in Chinese benchmarks, indicating strong competitiveness. We utilized BlueLM-7B-chat for our experiments.

\paragraph{GPT-3.5-Turbo}\footnote{\url{https://platform.openai.com/docs/models/gpt-3-5}} stands out as the most potent and cost-efficient model within the GPT-3.5 series. Tailored for conversations, it excels in comprehending and generating natural language. 

\paragraph{GPT-4}~\cite{openai_gpt-4_2023} is an advanced language model with enhanced understanding and generation capabilities. Trained on diverse internet text, it excels in various tasks, including text generation, translation, and problem-solving. We utilized GPT-4-1106-preview for our experiments.

\label{sec:Baselines}
\subsection{Implementation Details}
\label{app:Implementation Details}

\paragraph{Settings} When training TransferTOD-7B, we use Baichuan-7B-base as base model, formatting the data to adapt to the Baichuan training format. Training cost about 8 hours on 8 A800-80GB GPUs and some hyper-parameters of our training are shown in Table \ref{tab:hyperparameters}, each version of our TransferTOD-7B adopted the same hyper-parameters when training.\\
\begin{table}[ht]
    \centering
    \begin{tabular}{c|c}
    \toprule
    \textbf{HyperParameter} & \textbf{Value} \\
    \midrule
    num\_train\_epochs & 4 \\
    per\_device\_train\_batch\_size & 1 \\
    gradient\_accumulation\_steps & 1 \\
    learning\_rate & 9.65e-6 \\
    lr\_scheduler\_type & cosine \\
    adam\_beta1 & 0.9 \\
    adam\_beta2 & 0.98 \\
    adam\_epsilon & 1e-8 \\
    \bottomrule
    \end{tabular}
    \caption{Hyper-Parameters adopted when training TransferTOD-7B.}
    \label{tab:hyperparameters}
\end{table}

\paragraph{In-Domain test} When training in-domain models with dataset TransferTOD-v4, we tokenize the user utterance with Chinese pre-trained BERT~\cite{cui-etal-2021-pretrain} and annotate it with sequence labels using BIO tagging scheme.

\paragraph{Out-Of-Domain test} 
For the first part, evaluating the model's capability of slot filling. When inferencing with the LLMs in out-of-domain test, we meticulously designed a system prompt, describing the task and desired output format in detail, to get the best result from each LLM, while some chat models may still perform fairly bad for the slots in their output don't match JSON format. The system prompt used has been translated to English and showed in Table \ref{tab:prompt-zeroshot}.

For the second part, evaluating the semantic accuracy of model-generated questions, we use a manual evaluation approach. For detailed evaluation metrics, please refer to appendix~\ref{app:metrics}.

\begin{table}[ht]
    \centering
    \resizebox{\linewidth}{!}{
    \begin{tabular}{p{\linewidth}}
    \toprule
    \rowcolor{gray!10} \multicolumn{1}{c}{\textit{System}} \\
    You are an AI responsible for information extraction, and the scenario for information extraction is "<domain>". Based on your conversation with the user, please fill in the slots and continuously ask questions for the slots that are empty, with the number of slots to be asked in each question being <extract\_slot>. If the content of the user's answer includes information that does not belong to the slots you asked about in the previous round of conversation, do not fill in the slots with the incorrect parts of the user's answer. Instead, re-ask questions about the incorrect slots in the user's answer.\\\\
    
    The format of our input is as follows: Slots: \{"Slot\_1": "Value\_1", "Slot\_2": "Value\_2", ..., "Slot\_n": "Value\_n"\}\\
    The previous round of conversation: \{"assistant": "...", "human": "..."\}\\
    If there are still null slots after filling in, your response should follow this format: \{"Slot\_1": "Value\_1", "Slot\_2": "Value\_2", ..., "Slot\_n": "Value\_n"\}<Questions to ask>\\
    If there are no null slots after filling in, your response should follow this format: \{"Slot\_1": "Value\_1", "Slot\_2": "Value\_2", ..., "Slot\_n": "Value\_n"\} I have obtained all the information, and here is the content: \{"Slot\_1": "Value\_1", "Slot\_2": "Value\_2", ..., "Slot\_n": "Value\_n"\}\\
    \bottomrule
    \end{tabular}
    }
    \caption{The system prompt used prompting LLMs to execute out-of-domain test, where <domain> represents the domain of the test and <extract\_slot> represents the number of slots should be extracted in one turn.}
    \label{tab:prompt-zeroshot}
\end{table}

\subsection{Evaluation Metrics}
\label{app:metrics}
\paragraph{Joint Accuracy} measures the accuracy of dialogue states, considering a state correctly predicted only if all values of given slots are exactly matched.

Given the formula for Joint Accuracy is defined as:
\begin{equation}
JA = \frac{N_{cds}}{T_{ds}}
\end{equation}
where $JA$ denotes \textbf{Joint Accuracy},$N_{cds}$ stands for the \textbf{Number of dialog states correctly predicted},and $T_{ds}$ represents the \textbf{Total number of dialog states}.

\paragraph{Slot F1} calculates the F1 score of (slot, value) pairs, deeming a tuple correctly predicted if the slot's value is exactly matched.\\
Given the formula for Slot F1 is defined as:



\begin{equation}
    \textbf{SlotF1} = \frac{1}{N_{Slots}}\sum_{i=1}^{N_{Slots}}\textbf{F1 Score}_i
\end{equation}
where $N_{Slots}$ represents the \textbf{total number of (slot, value) pairs}.

\paragraph{Dialogue Act F1} calculates the F1 score of (intent, slot, value) dialogue acts, where intent are always "inform", deeming a dialogue act correctly predicted if the slot and value extracted from user utterance is exactly matched.
Given the formula for Dialogue Act F1 is defined as:
\begin{equation}
    \textbf{Dialogue Act F1} = \frac{\sum_{i=1}^{N_{DialogueActs}}\textbf{F1 Score}_i}{N_{DialogueActs}}
\end{equation}
where $N_{DialogueActs}$ represents the \textbf{total number of (intent, slot, value) dialogue acts}.

\paragraph{Ask Accuracy} measures the model's ability to correctly select the corresponding number of slots from empty slots or to correctly point out errors in user answers and ask questions that correspond to the correct slots and will not cause misunderstandings. 

\begin{equation}
    \textbf{Ask Accuracy} = \frac{\sum_{i=0}^{3} i \times A_i}{N \times 3} \times 100
\end{equation}
where $A_{i}$ represents the \textbf{number of the dialogues that got a score of accuracy i which ranks from 0 to 3} and $N$ represents the \textbf{total number of the dialogues}.

For question accuracy scores, the scoring rules are as follows:\\
- 0 points represent that the model's questions are ambiguous, or it fails to correctly select fields from the empty slots for questioning, and the number of questioned fields does not match \{extract\_slot\} (if the number of remaining empty fields is less than \{extract\_slot\} and the number of questions asked does not equal the total of all remaining empty fields while meeting the previous condition, it should also be categorized here).\\
- 1 point represent that the model's questions might cause ambiguity, but it can correctly select fields from the empty slots for questioning, yet the number of questioned fields does not match \{extract\_slot\} (if the number of remaining empty fields is less than \{extract\_slot\}) and the number of questions asked does not equal the total of all remaining empty fields while meeting the first two conditions, it should also be categorized here).\\
- 2 points represent that the model's questions are precise, unambiguous, and it can correctly select fields from the empty slots for questioning, but the number of questioned fields does not match \{extract\_slot\}) (if the number of remaining empty fields is less than \{extract\_slot\} and the number of questions asked does not equal the total of all remaining empty fields while meeting the first two conditions, it should also be categorized here).\\
- 3 points represent that the model's questions are precise, unambiguous, and it can correctly select fields from the empty slots for questioning, and the number of questioned fields matches \{extract\_slot\}) (if the number of remaining empty fields is less than \{extract\_slot\}) and the number of questions asked equals the total of all remaining empty fields while meeting the first two conditions, it should also be categorized here). If all slots are filled and the model does not initiate a question or says "I have obtained all the information," the message content "" should also fall into this category.

\paragraph{Ask Fluency} measures the fluency of the model’s questions and the degree to which they are consistent with natural language features. 
\begin{equation}
    \textbf{Ask Fluency} = \frac{\sum_{i=0}^{3} i \times F_i}{N \times 3} \times 100
    \label{eq:fluency_score}
\end{equation}
where $F_{i}$ represents the \textbf{number of the dialogues that got a score of fluency i whick ranks from 0 to 3} and $N$ represents the \textbf{total number of the dialogues}.

For the fluency score, experts rate the model's questions on fluency across a scale of 0 to 3 points. \\
- 0 points represent that the representative's questioning style is rigid and awkward, completely deviating from the characteristics of natural language.\\
- 1 point represent that the representative's questioning style is somewhat rigid, yet the language is relatively natural, aligning with certain characteristics of natural language.\\
- 2 points represent that the representative's questioning style is relatively natural, and the language used is also quite consistent with the characteristics of natural language.\\
- 3 points represent that the representative's questioning style is very natural, and the language fully complies with the characteristics of natural language.\\

We referred to the work of LLMEval \cite{zhang2024llmeval} to design the evaluation criteria for accuracy and fluency.

\section{Templates for script-generated data}
\label{app:templated}
Table \ref{tab:script-generated-template} shows an example of a question and answer template when Domain set to ‘Hotel' and Slot set to ‘Hotel Type'.

\begin{table}[ht]
    \centering
    \resizebox{\linewidth}{!}{
    \begin{tabular}{p{\linewidth}}
    \toprule
    \rowcolor{gray!10} \multicolumn{1}{c}{\textit{Question template}} \\
    "Hotel Type": [ \\
    ~"What type of hotel do you prefer? For example, luxury hotels, economy hotels, etc.", \\
    ~"Do you have any specific preferences for hotel types?", \\
    ~"Please tell us the type of hotel you'd like to stay in, such as resorts, city hotels, etc." \\
    ] \\
    \rowcolor{gray!10} \multicolumn{1}{c}{\textit{Answer template}} \\
    "Hotel Type": [ \\
    ~"We would like to stay at {} hotel.", \\
    ~"We want {} hotel.", \\
    ~"We wish to stay at {} hotel.", \\
    ~"I prefer {} hotel.", \\
    ~"I particularly like {} hotel.", \\
    ~"I hope to stay at {} hotel." \\
    ] \\
    \bottomrule
    \end{tabular}
    }
    \caption{An example of a question and answer template.}
    \label{tab:script-generated-template}
\end{table}

\section{Prompts}
\label{prompts}
\subsection{Prompt GPT-3.5 to Polish the Data}
The prompt showed in Table \ref{tab:prompt-polish_question} and Table \ref{tab:prompt-polish_answer} are used when using GPT-3.5 to polish the text, rewriting questions and answers respectively, in our dataset.

\begin{table}[ht]
    \centering
    \resizebox{\linewidth}{!}{
    \begin{tabular}{p{\linewidth}}
    \toprule
    \rowcolor{gray!10} \multicolumn{1}{c}{\textit{User}} \\
    You are a \{domain\} company front desk customer service. The following content is the question you want to ask the user. Please change the wording to ask the question. You do not need to output other content, you only need to complete the rewriting.\\\\
    Original question: \{question\}\\
    Here's a rephrased version of your question:\\
    \bottomrule
    \end{tabular}
    }
    \caption{The prompt for rewriting the question.}
    \label{tab:prompt-polish_question}
\end{table}

\begin{table}[ht]
    \centering
    \resizebox{\linewidth}{!}{
    \begin{tabular}{p{\linewidth}}
    \toprule
    \rowcolor{gray!10} \multicolumn{1}{c}{\textit{User}} \\
    You are a user, the following is the original answer, the specific content name can not be changed, such as the level, service name, etc., please answer in a different expression. You do not need to output other content, just complete the rewrite.\\\\
    Original answer: \{answer\}\\
    Here is your answer with a different formulation:\\
    \bottomrule
    \end{tabular}
    }
    \caption{The prompt for rewriting the answer.}
    \label{tab:prompt-polish_answer}
\end{table}

\subsection{Prompt GPT-4 to Evaluate the Results}
\label{app:Prompt GPT-4 to evaluate the results}

The prompt in Table \ref{tab:prompt-gpt-diversity} is used when using GPT-4 to conduct comparative evaluation of diversity in ablation experiments, while the prompt in Table \ref{tab:prompt-gpt-fluency} is used when using GPT-4 to score the fluency (Ask\_Flu) of model questions in ablation experiments.

\begin{table*}[!t]
    \centering
    \resizebox{\linewidth}{!}{
    \begin{tabular}{p{\linewidth}}
    \toprule
   \rowcolor{gray!10} \multicolumn{1}{c}{\textit{User}} \\
The following is a dialogue scenario for a task of information extraction, where two customer service representatives are inquiring customer information. You are required to compare the diversity in questioning styles and sentences between two groups in order to evaluate their performance.\\\\

Your options are as follows:\\
Option A: Group A's questioning style is noticeably more diverse than Group B's.\\
Option B: Group B's questioning style is noticeably more diverse than Group A's.\\
Option C: Both Group A and Group B demonstrate a similar level of diversity in their questioning.\\
Option D: Both Group A and Group B lack diversity in their questioning.\\\\

The inquiries from customer service A are as follows: \{selected\_a\_questions\}\\
The inquiries from customer service B are as follows: \{selected\_b\_questions\}\\\\

You must provide your feedback in the following format:\\
Reason: reason\\
Option: A, B, C or D\\
    \bottomrule
    \end{tabular}
    }
    \caption{The prompt for using GPT-4 to conduct comparative evaluation of diversity in ablation experiments.}
    \label{tab:prompt-gpt-diversity}
\end{table*}










\begin{table*}[!t]
    \centering
    \resizebox{\linewidth}{!}{
    \begin{tabular}{p{\linewidth}}
    \toprule
   \rowcolor{gray!10} \multicolumn{1}{c}{\textit{User}} \\
The following scenario is a customer service question asked by a user to obtain specific information. You need to rate the fluency of the customer service question. Fluency includes factors such as whether the question is a complete sentence, whether it contains pauses of unclear meaning, whether the questioning method is blunt, whether it conforms to the characteristics of natural language, etc., and customer service questions are scored accordingly. If the customer service says "I have obtained all the information, the following is the information content" and is followed by a json string, the item will be rated as a full score.\\\\

\textbf{Fluency:}\\
- 0 points mean that the customer service's questions are not fluent. Multiple questions are divided into many independent questions, or contain pauses with unclear meaning. The questioning method is stiff. Completely inconsistent with the characteristics of natural language\\
- 1 point means that the customer service questions are not fluent. Multiple questions are divided into multiple short sentences, or contain relatively abrupt pauses. Not consistent with the characteristics of natural language\\
- 2 points mean that the customer service questions are relatively fluent, and multiple questions are relatively fluently combined into long sentences, which is more in line with the characteristics of natural language.\\
- 3 points mean that the customer service questions are very fluent, and multiple questions are fluently combined into long sentences, which fully conforms to the characteristics of natural language.\\\\

The customer service question content is as follows: \{ques\}\\\\

You must give your feedback in the following format:\\
Reason: reason\\
Fluency: score of its fluency (int)\\
    \bottomrule
    \end{tabular}
    }
    \caption{The prompt for using GPT-4 to score the fluency (Ask\_Flu) of model's questions in ablation experiments.}
    \label{tab:prompt-gpt-fluency}
\end{table*}

\section{Data Examples}
\label{data_examples}
\begin{figure*}
    \includegraphics[width=0.7\linewidth]{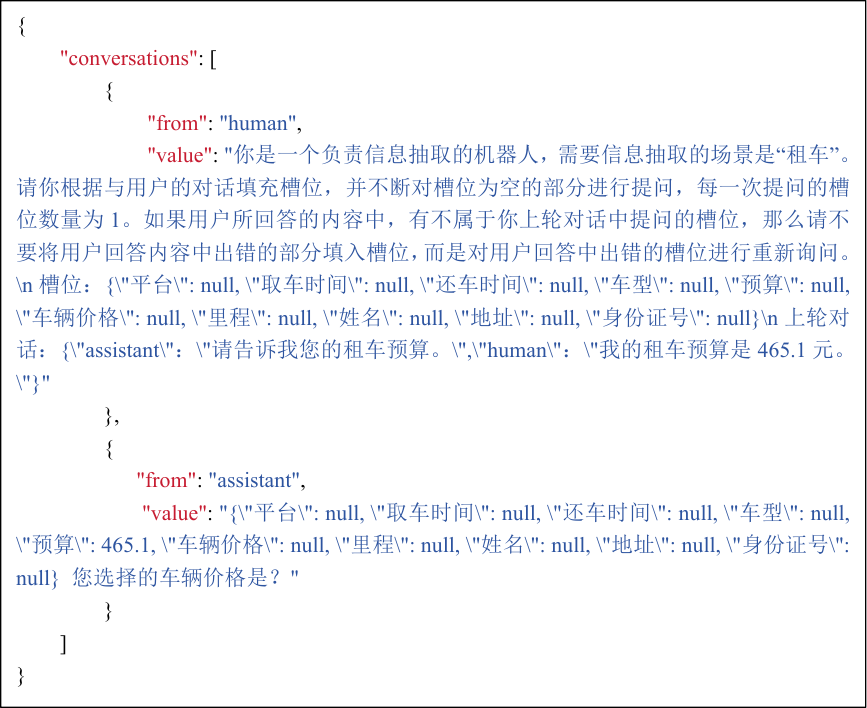}
    \centering
 	\caption{
An example of the training data for supervised-finetuning TransferTOD-7B
  }
	\label{fig:sft_data-ch}
\end{figure*}

\begin{figure*}
    \includegraphics[width=0.7\linewidth]{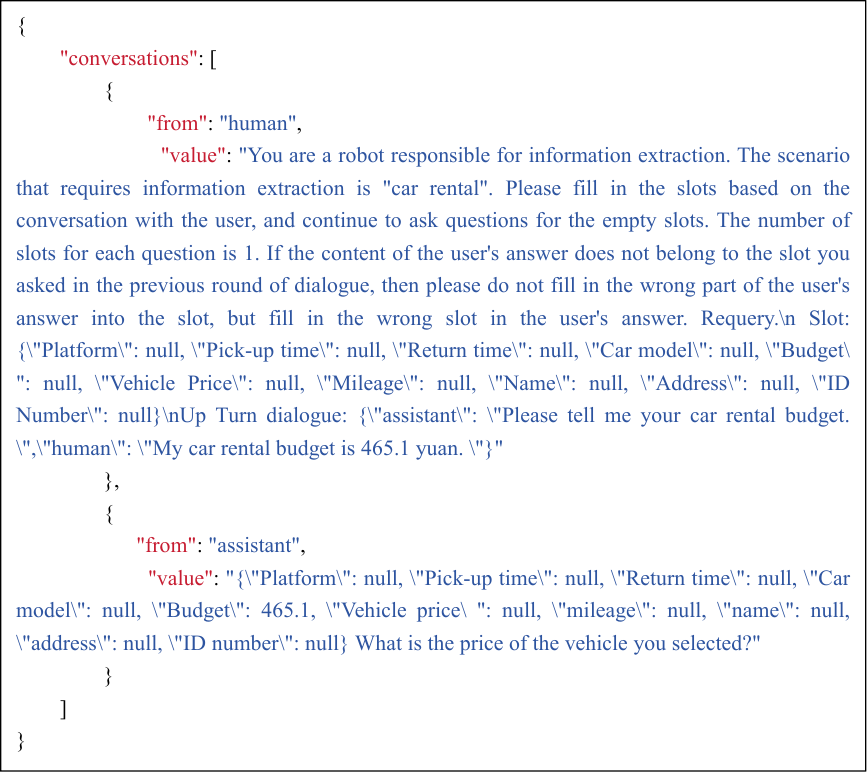}
    \centering
 	\caption{
An example of the training data for supervised-finetuning TransferTOD-7B (in English)
  }
	\label{fig:sft_data-en}
\end{figure*}

\begin{figure*}
    \includegraphics[width=0.7\linewidth]{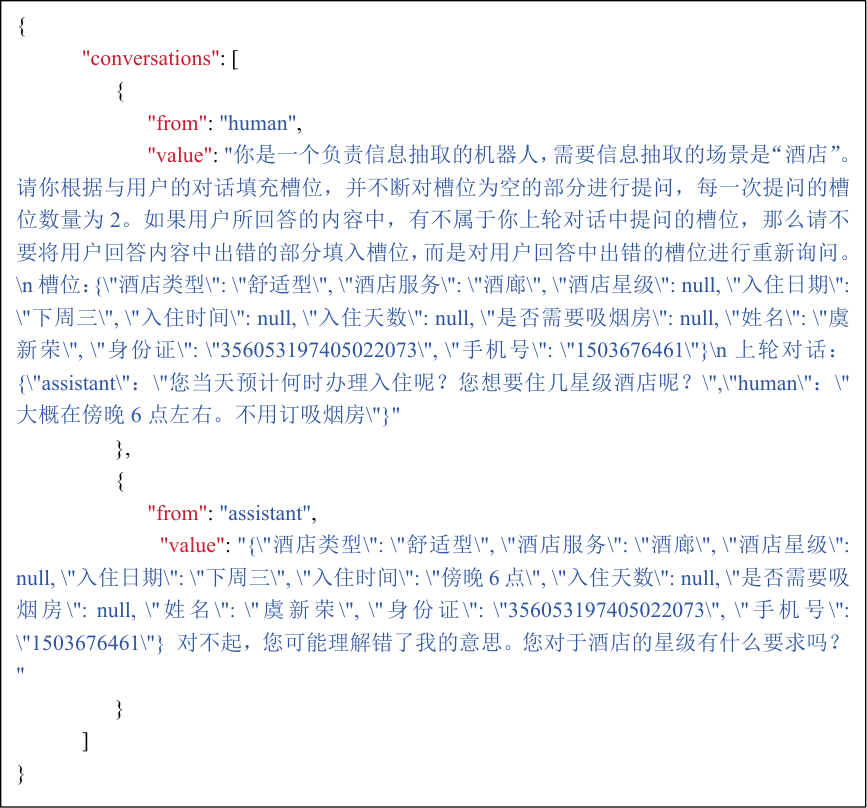}
    \centering
 	\caption{
An example of the training data with noise for supervised-finetuning TransferTOD-7B
  }
	\label{fig:sft_noise_data-ch}
\end{figure*}

\begin{figure*}
    \includegraphics[width=0.7\linewidth]{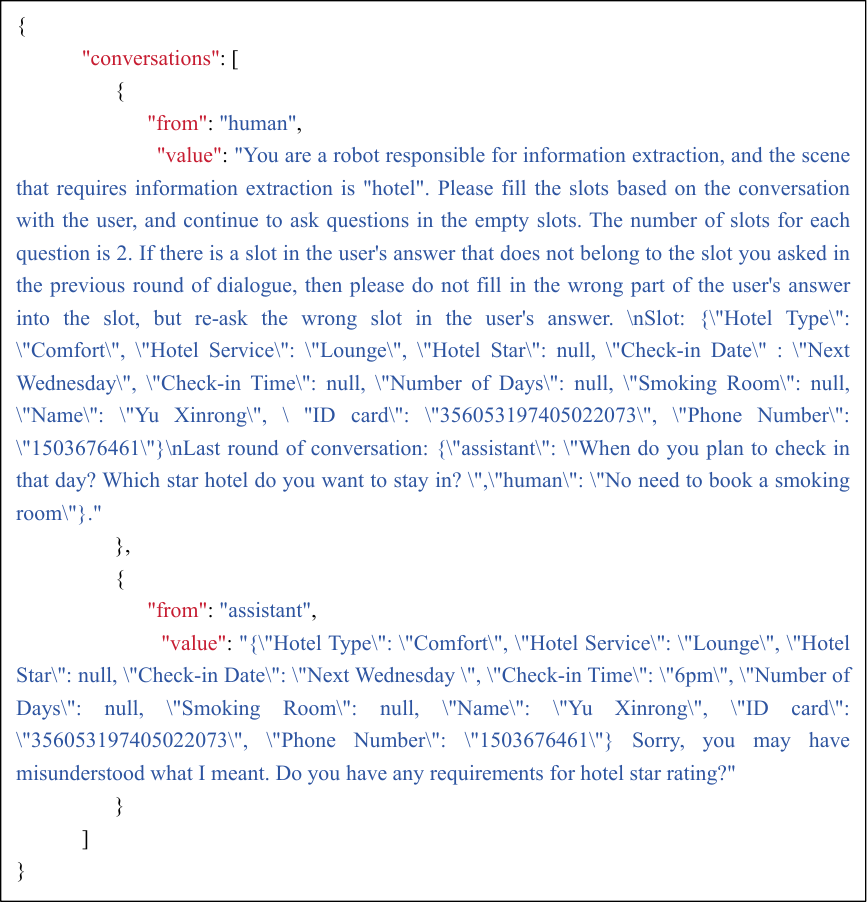}
    \centering
 	\caption{
An example of the training data with noise for supervised-finetuning TransferTOD-7B (in English)
  }
	\label{fig:sft_noise_data-en}
\end{figure*}

\begin{figure*}
    \includegraphics[width=0.85\linewidth]{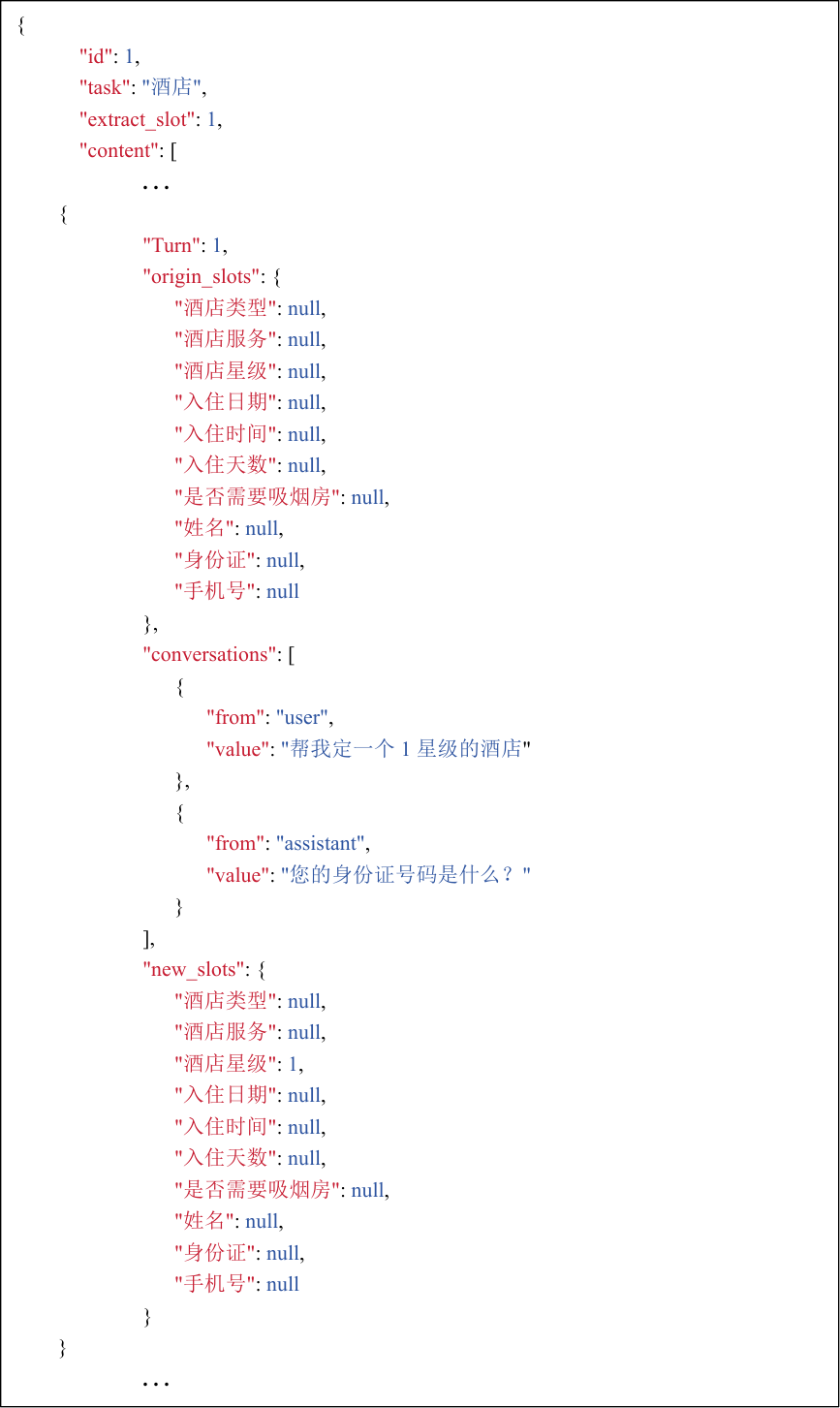}
    \centering
 	\caption{
An example of our TransferTOD dataset
  }
	\label{fig:raw_data-ch}
\end{figure*}

\begin{figure*}
    \includegraphics[width=0.85\linewidth]{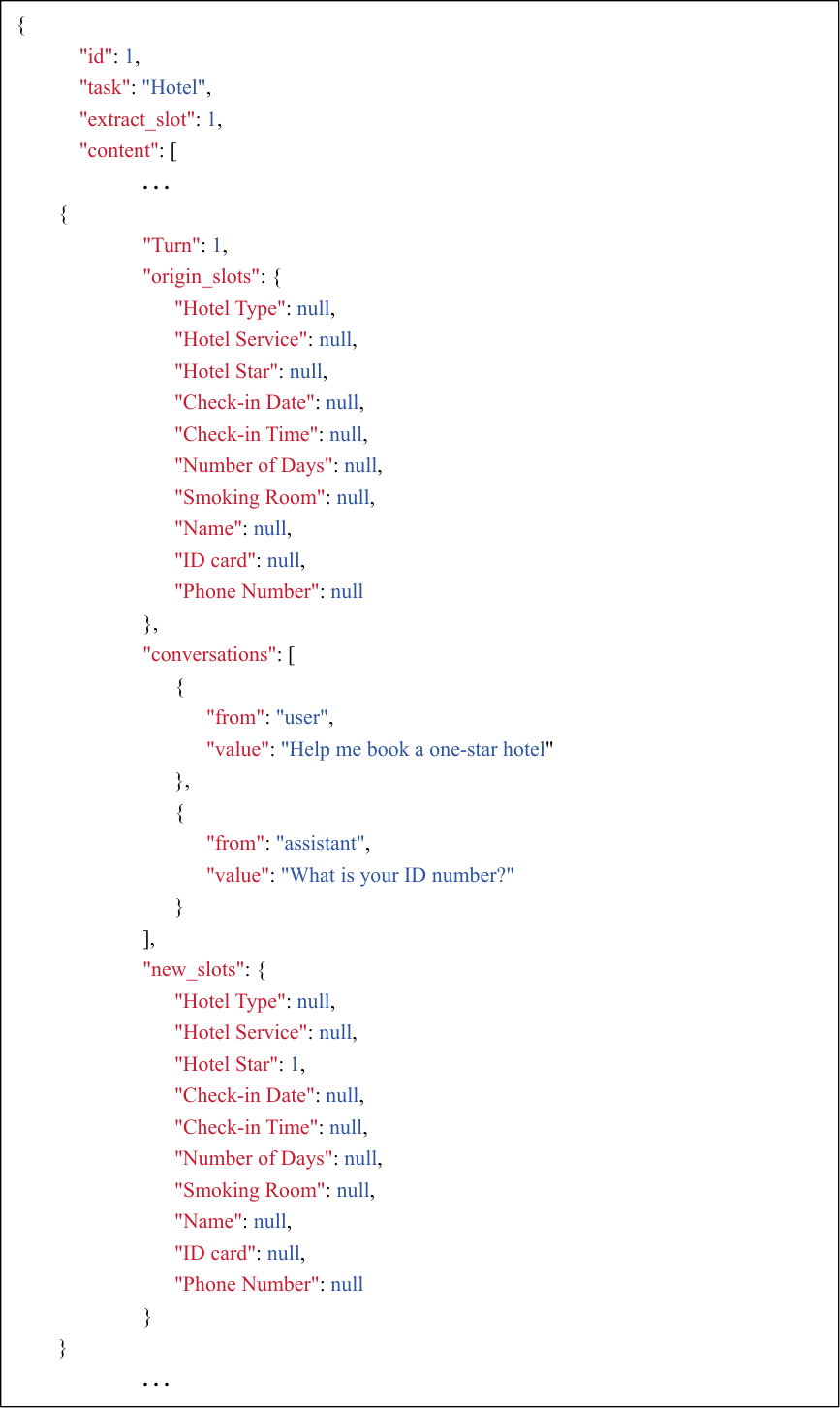}
    \centering
 	\caption{
An example of our TransferTOD dataset (in English)
  }
	\label{fig:raw_data-en}
\end{figure*}

Examples of our supervised-finetuning data are showed in Figure \ref{fig:sft_data-ch} and Figure \ref{fig:sft_data-en}, also we provide examples of data with noise in Figure \ref{fig:sft_noise_data-ch} and Figure \ref{fig:sft_noise_data-en} as well as raw TransferTOD data in Figure \ref{fig:raw_data-ch} and Figure \ref{fig:raw_data-en}.

\section{Human Experts}
\subsection{Experts in Constructing Datasets}
\label{app:Experts in Constructing Datasets}

During the dataset construction phase, we relied on 5 students from our institute to participate in this work as human experts. These students possessed good computer knowledge and coding skills, which enabled them to perform the task effectively.

Their primary responsibility is to generate dialogue data for test sets. We assign tasks based on different scenarios, ensuring they are familiar with the entire dataset construction process and principles. They work professionally, providing human support for the dataset creation and ensuring smooth project execution. Additionally, we fairly compensate their efforts to show respect and recognition for their contributions.

 Another task for human experts involves refining non-fluent content. Given the potential for incoherence and unnaturalness in rule-based generation in \ref{sec:Dialog construction}, characterized by the lack of appropriate connective words and inconsistent tone, we prioritize addressing this issue.Thus, human experts are employed to revise dialogue content, such as transforming "What's your name? Please tell me your phone number." into a more coherent and natural structure like "Please provide your name and phone number."

Compared to rule-based mass generation, expert-crafted data exhibits significant advantages. The work of domain experts enhances the linguistic fluency, naturalness, and brevity of the generated dialogues. This high-quality, manually constructed data boasts greater authenticity and representativeness, more effectively emulating real-world conversation scenarios. Consequently, it serves as a more reliable foundation for subsequent fine-tuning tasks.

\subsection{Experts in Ablation Experiment}
\label{app:Experts in Ablation Experiment}
During the ablation experiment phase, we invited 12 students from our institution to conduct comparative evaluations of the results. Each student was assigned to complete the full assessment tasks for one or more large models. This entailed each student conducting a comprehensive evaluation of the designated model to ensure a thorough understanding of its performance.

Specifically, we selected 200 data points from the inference results of TransferToD-7B-v2 and TransferToD-7B-v3, and conducted 200 random samples. 5 data points were sampled each time, resulting in a total of 40 evaluations for each model's inference results. This random sampling method contributed to ensuring the objectivity and reliability of the assessment, minimizing potential biases.

Subsequently, the evaluators rated the sampled data based on the questioning style, diversity, and fluency. They provided an overall score for each set of data by considering factors such as the model's questioning approach, sentence completeness, clarity of questioning, diversity, and fluency. These scores provided quantitative data on the model's performance in various aspects, facilitating a more comprehensive assessment and comparison of the models' strengths and weaknesses.

\begin{table*}
    \centering
    \resizebox{\linewidth}{!}{
    \begin{tabular}{c  |c c c c c }
    \toprule
     {\textbf{Model}}& \textbf{Scenario} & \textbf{JointAcc(\%)} & \textbf{SlotF1(\%)} & \textbf{AVG.JointAcc(\%)} & \textbf{AVG.SlotF1(\%)} \\
    \midrule
    \multirow{3}{*}{\textbf{TransferTOD-7B}} & Water-Delivery & \textbf{75.16}& \textbf{96.61}& \multirow{3}{*}{\textbf{75.09}}& \multirow{3}{*}{\textbf{96.20}}\\
      & Sanitation& \textbf{84.09}& \textbf{97.43}& & \\
      & Courier& \textbf{68.00}& \textbf{94.57}& & \\
     \cmidrule{1-6}
      \multirow{3}{*}{\textbf{Baichuan2-7B-Chat(5-shot)}}& Water-Delivery & 52.40& 82.16& \multirow{3}{*}{53.78}& \multirow{3}{*}{82.42}\\
      & Sanitation& 71.71& 94.92& & \\
      & Courier& 37.23& 70.19& & \\
     \cmidrule{1-6}
      \multirow{3}{*}{\textbf{BlueLM-7B-Chat(5-shot)}}& Water-Delivery & 61.98& 93.87& \multirow{3}{*}{42.81}& \multirow{3}{*}{86.32}\\
      & Sanitation& 43.90& 87.54& & \\
      & Courier& 22.54& 77.57& & \\
     \cmidrule{1-6}
      \multirow{3}{*}{\textbf{Chatglm3-6B(5-shot)}}& Water-Delivery & 22.92& 53.35& \multirow{3}{*}{27.32}& \multirow{3}{*}{64.24}\\
      & Sanitation& 31.43& 67.42& & \\
      & Courier& 27.62& 71.96& & \\
     \cmidrule{1-6}
      \multirow{3}{*}{\textbf{Qwen-7B-Chat(5-shot)}}& Water-Delivery & 69.09& 94.04& \multirow{3}{*}{61.69}& \multirow{3}{*}{91.44}\\
      & Sanitation& 61.14& 91.26& & \\
      & Courier& 54.85& 89.02& & \\
     \cmidrule{1-6}
      \multirow{3}{*}{\textbf{Yi-6B-Chat(5-shot)}}& Water-Delivery & 67.89& 94.94& \multirow{3}{*}{63.09}& \multirow{3}{*}{94.04}\\
      & Sanitation& 64.00& 93.87& & \\
      & Courier& 57.38& 93.32& & \\
      \cmidrule{1-6}
      \multirow{3}{*}{\textbf{GPT-4-1106-Preview(5-shot)}}& Water-Delivery & 65.10& 75.98& \multirow{3}{*}{65.39}& \multirow{3}{*}{76.47}\\
      & Sanitation& 65.14& 75.87& & \\
      & Courier& 65.92& 77.57& & \\
     \bottomrule
    \end{tabular}
    }
    
    \caption{Result of out-of-domain test with the setting of In-Context Learning: The Joint Accuracy and Slot F1 Score of each model, showing the accuracy of predicting the right dialogue state and slot-value pairs respectively.}
    \label{tab:ood_result_icl}
\end{table*}

\end{document}